\documentclass[letterpaper, 10 pt, conference]{ieeeconf}  % Comment this line out if you need a4paper

\IEEEoverridecommandlockouts                              % This command is only needed if 
\usepackage{comment}
\usepackage{algpseudocode}
\usepackage{xcolor}
\usepackage{multirow}

\usepackage{times}
\usepackage{epsfig}
\usepackage{amsmath}
\usepackage{amssymb}

\usepackage{siunitx}

\usepackage{graphicx}
\usepackage{booktabs}
\usepackage{algorithm}
\usepackage{cite}

\usepackage[font=small,labelfont=bf]{caption}
\usepackage{subcaption}
\usepackage{caption, subcaption}
\usepackage[switch]{lineno}

\usepackage[pagebackref=true,breaklinks=true,colorlinks,bookmarks=false,citecolor=red]{hyperref}

\title{MapFusion: A General Framework for 3D Object Detection \\with HDMaps
}

\author{Jin Fang$^{1, 2}$, Dingfu Zhou$^{1, 2}$, Xibin Song$^{1, 2}$, Liangjun Zhang$^{1, 2}$

\thanks{$^{1}$ Robotics and Autonomous Driving Laboratory, Baidu Research.  $^{2}$ National Engineering Laboratory of Deep Learning Technology and Application, China.}
\thanks{
        {\tt\small \{fangjin, zhoudingfu, songxibin, liangjun zhang\}@baidu.com}
        }%    
}

\begin{document}

\maketitle
\thispagestyle{empty}
\pagestyle{empty}

%%%%%%%%%%%%%%%%%%%%%%%%%%%%%%%%%%%%%%%%%%%%%%%%%%%%%%%%%%%%%%%%%%%%%%%%%%%%%%%%
\begin{abstract}
% 3D object detection is critically important for autonomous driving. The most recent proposed approaches are based on Lidar sensor only or fused with cameras. Although the performances are relatively satisfactory, there is space for improvement. In addition, Map (e.g., High Definition Maps), as a basic infrastructure for intelligent vehicles, has not been well explored for booting object detection tasks. Here, we propose a simple but effective framework-MapFusion to integrate the map information into the modern 3D object detectors. In particular, the proposed MapFusion is detector independent and can be easily merged into different detectors. The experimental results of three different baselines on the large public autonomous driving dataset demonstrate the superiority of the proposed framework. By adding the map information, we can achieve xxxx  %2.79 point improvements for mean Average Precision (mAP).

3D object detection is a key perception component in autonomous driving. Most recent approaches are based on Lidar sensors only or fused with cameras. Maps (e.g., High Definition Maps), a basic infrastructure for intelligent vehicles, however, have not been well exploited for boosting object detection tasks. In this paper, we propose a simple but effective framework - MapFusion to integrate the map information into modern 3D object detector pipelines. In particular, we design a FeatureAgg module for HD Map feature extraction and fusion, and a MapSeg module as an auxiliary segmentation head for the detection backbone. Our proposed MapFusion is detector independent and can be easily integrated into different detectors. The experimental results of three different baselines on large public autonomous driving dataset demonstrate the superiority of the proposed framework. By fusing the map information, we can achieve 1.27 to 2.79 points improvements for mean Average Precision (mAP) on three strong 3d object detection baselines.
%respectively.

 %can be improved we can achieve xxxx point improvements for mean Average Precision (mAP).

%HDMap (High Definition Map) information is critical for autonomous driving, due to its high precision and contribution to computation reduction. But for perception module, HDMap is usually used as post-processing. We propose a general MapFusion framework, which contributes to the detection performance by integrating the map information into the detection framework, and improve the feature representation ability of point cloud. 

\end{abstract}

\section{Introduction}\label{sec:intro}

%WE NEED TO REWRITE THE FIRST FEW SENTENCES. 

Autonomous driving (AD) has drawn significant attention over the past years. Despite recent progress, AD is still considered as one of the most challenging tasks \cite{Takeda2020Survey}. Developing autonomous driving systems needs to integrate many advanced robotics techniques such as perception, planning, and control. Perception module (e.g., object detection \cite{arnold2019survey}, lane-marks segmentation,
% \cite{ADD one paper for segmentation}
traffic lights detection, etc.) is particularly important in the AD system. Perception module interprets the surrounding environment and provides inputs for downstream planning and control modules. For object detection tasks, cameras are the most commonly used sensors and many advanced camera-based 2D detectors \cite{ren2015faster,liu2016ssd}
% such as Faster R-CNN \cite{ren2015faster} and SSD \cite{liu2016ssd} 
have been developed. However, autonomous driving vehicles need to recover 3D locations of surrounding objects accurately and reliably, which are extremely hard for 2D image-based detectors, due to the loss of distance information in the perspective projection process from 3D space to 2D image. Therefore, active 3D sensors such as LiDAR devices, are prerequisites for most AD systems.

\begin{figure}[t!]
	\centering
	\includegraphics[width=0.49\textwidth]{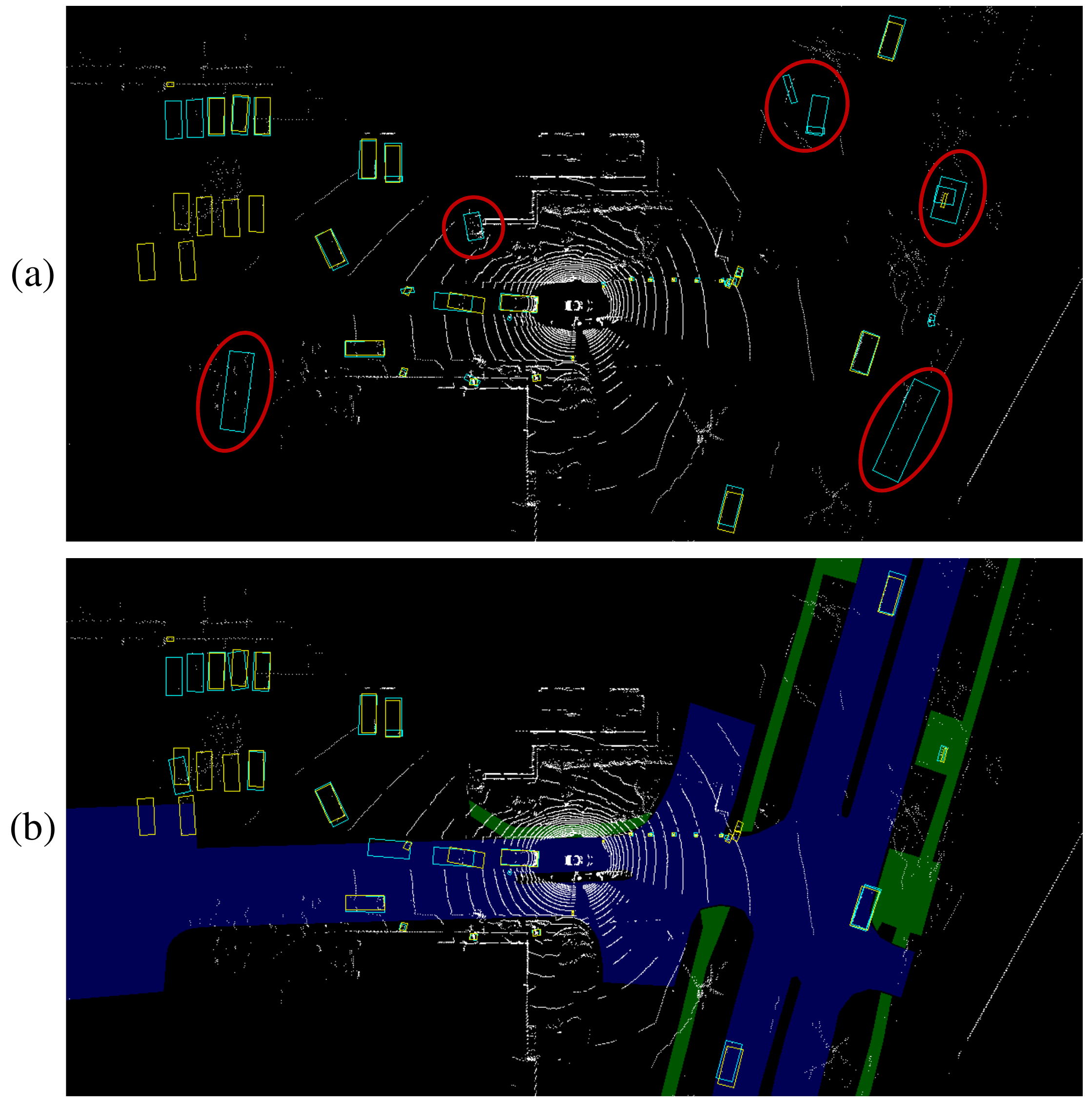}
	\centering
	\caption{(a) shows the detection results from PointPillars\cite{lang2019pointpillars}, where the false positives are marked by red cycles. (b) shows the detection results from PointPillars with MapFusion, where the false positives are removed benefiting from the MapFusion framework. The yellow rectangles are the ground truth and the cyan rectangles are the predict results.}
	\label{Fig:detect_compare}
\end{figure}

%The false positives marked by red cycle in (a) are removed benefiting from our MapFusion framework.

With the development of deep learning on the 3D point clouds, many LiDAR-based 3D object detectors have been proposed. Based on the different input representations, approaches can be generally categorized as point-based \cite{shi2019pointrcnn, zhou2020joint} and voxel-based methods \cite{lang2019pointpillars, yan2018second}. Compared with point-based methods, the latter is much more efficient and its computation time is independent of the size of the point cloud. The performance for existing LiDAR-based 3D object detection approaches, however, can be further improved, especially on reducing the false positives and false negatives for detected objects. This problem comes from two main reasons. First, without enough texture information, it is difficult to distinguish between foreground and background objects. Second, the number of scanned points for small objects or objects far away is very sparse, which results in the failure of object detection and recognition. 
%(\liangjun{HOW SPARSE POINTS results in false positive. If there are too few points, we may not be able to detect them. This would be false negative}) . 
Fusing with other sensors e.g., camera \cite{liang2019multi}, radar \cite{wang2021rodnet} is a commonly used strategy to handle this kind of problem. However, as image quality is easily affected by the environment illumination and weather situation, solely relying on fusion with images can't provide stable detection results. Fig.~\ref{Fig:detect_compare} (a) demonstrates the 3D detection results of PointPillars\cite{lang2019pointpillars}, and we can see that false positives occurs due to the problems discussed above.

% With the development of deep learning on the 3D point clouds, many LiDAR-based 3D object detectors have been proposed. Based on the different input representations, approaches can be generally categorized as point-based \cite{shi2019pointrcnn, zhou2020joint} and voxel-based methods \cite{lang2019pointpillars, yan2018second}. Compared with point-based methods, the latter is much more efficient and its computation time will not change too much with the increasing of thblem comes from two main reasons. First, without enough texture information, it is really difficult to distinguish between foreground object and backe point cloud number. With only the Lidar sensor, most of the approaches have achieved satisfactory performance except for many false positives. This proground. Second, the scanned points from the far object are very small which is also a big issue for object detection and recognition. Fused with other sensors e.g., camera \cite{liang2019multi}, radar \cite{wang2021rodnet} is a commonly used strategy to handle this kind of problem. However, the image quality is easily affected by the environment illumination and weather situation and can't provide stable detection results in extreme conditions.  

 \begin{figure*}[ht!]
	\centering
	\includegraphics[width=0.85\textwidth]{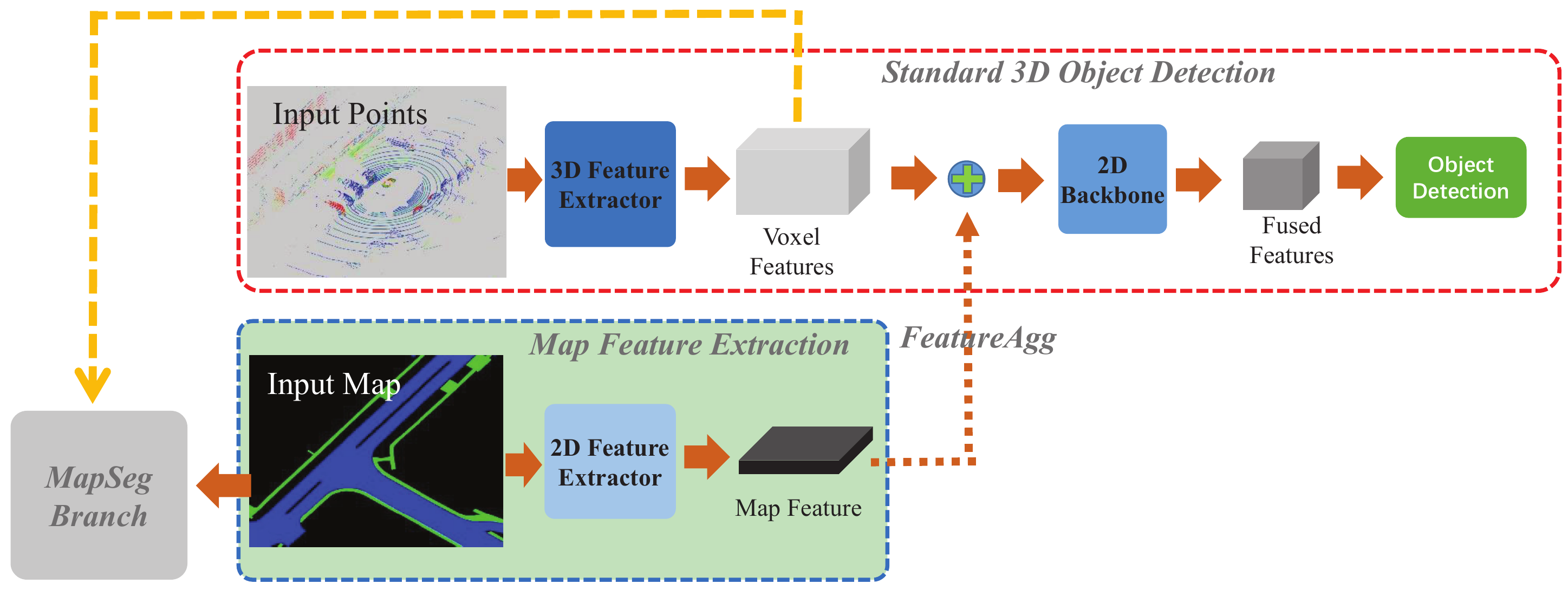}
	\centering
	\caption{
	The overview of our proposed MapFusion framework. The modules in the red dotted box are standard 3D object detection and the elements in the blue dotted box are our map feature extraction block. The extracted map features are concatenated with the voxel features for the following object detection branch. In addition, a sub-module called MapSeg branch is followed the voxel features for predicting the road map from the input point cloud, which is supervised by the ground truth map.
	} 
	\label{Fig:Framework}
\end{figure*}

In this paper, we exploit boosting the performance of perception tasks such as 3D object detection using High Definition Map (HDMap). Different from regular GPS digital maps, which are primarily meant for human navigation, HDMaps specifically have extremely high precision at centimeter-level and include all topological and geometric information of roads, such as lanes, traffic signs, and sidewalk zones. Most existing 3D object detection approaches only use the HDMap during the post-processing procedure to remove false positive detection such as objects far from the roads. Few approaches have been proposed to integrate the HDMap information into the detection network to improve the performance. In HDNET \cite{yang2018hdnet}, a self-designed 3D object detection pipeline has been designed for using HDMaps as inputs, though their detection method is based on the bird-eye-view representation by projecting the 3D point cloud into 2D. 

% Intuitively, High Definition Map (HDMap) which includes all the topological and geometric information of roads, such as lanes, traffic signs, and sidewalk zones, can be very useful for perception tasks. However, most existing 3D object detection approaches only use the HDMap during the post-processing procedure to remove false positive detection which are from the roads. Few approaches have been proposed to integrate the HDMap information into the detection network to boost the performance. In HDNET \cite{yang2018hdnet}, a self-designed 3D object detection pipeline has been designed for using HDMaps as inputs, though their detection method is based on the bird-eye-view representation by projecting the 3D point cloud into 2D. 

In this work, we propose a general fusion framework for integrating the HDMap information to boost the 3D object detection task. We design a FeatureAgg module for HDMap feature extraction and fusion, and a MapSeg module as an auxiliary segmentation head for the detection backbone. More importantly, the designed fusion framework is detector independent and can be employed for different 3D object detection pipelines directly, such as SECOND \cite{yan2018second}, PointPilars \cite{lang2019pointpillars} and CenterPoint \cite{yin2020center}. The effectiveness of the proposed framework has been verified on the large-scale public autonomous driving dataset-nuScenes \cite{nuscenes2019}. As shown in Fig.~\ref{Fig:detect_compare} (b), false positives can be partly 
alleviated, which proves the effectiveness of our approach.

The contributions of the proposed work can be generally summarized as:

\begin{enumerate}
\item We present MapFusion, a general data fusion framework for 3D object detection in AD scenario, which is more general, effective, and can be integrated into any 3D detection pipelines. 
\item A light-weight network has been designed to extract useful features from the HDMaps and the overhead for inference time has only and false negatives increased slightly compared with the baselines.
\item We evaluate the proposed framework on large-scale public dataset nuScenes and the experimental results demonstrate that the map information can significantly improve the performance of 3d object detection.
\end{enumerate}

\section{Related Work}\label{sec:related_work}

\begin{figure*}[ht!]
	\centering
	\includegraphics[width=0.7\textwidth]{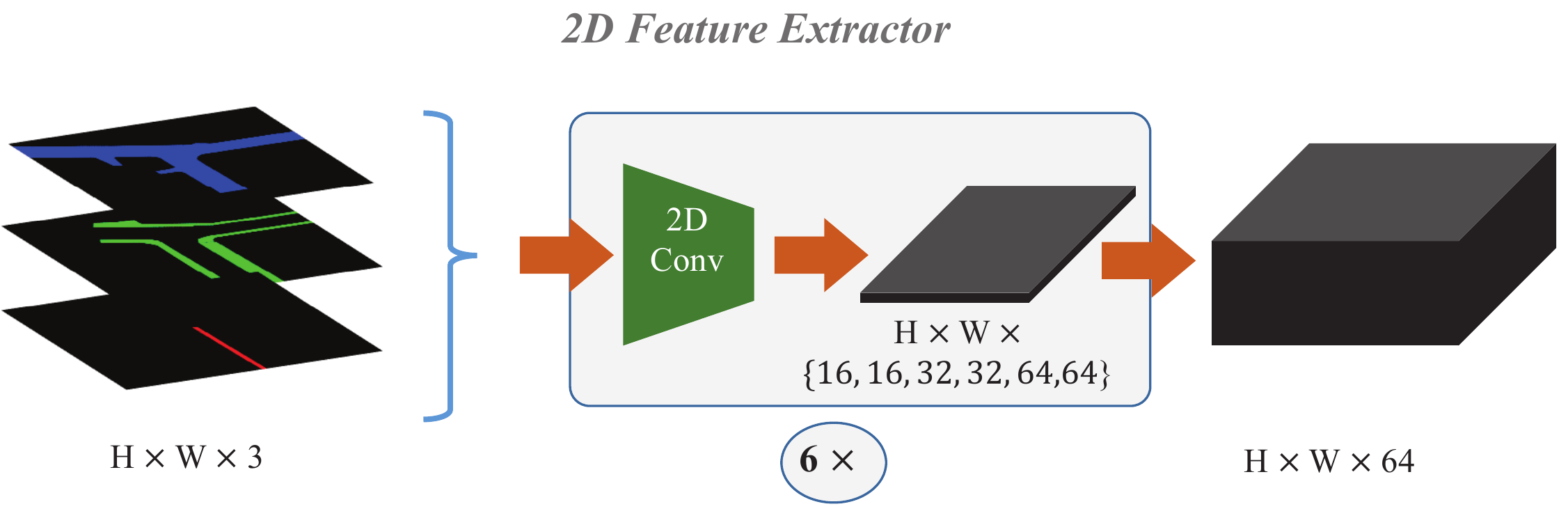}
	\centering
	\caption{The architecture of the \textit{2D Feature Extractor} contains 6 similar layers. Each layer contains a 2D convolution operation, a batch normalization operation and a ReLU function.}
	\label{Fig:2d_feature_extractor}
\end{figure*}

\subsection{Lidar-based 3d Object Detection}

With the development of range sensors and AD techniques, 3D object detection in driving scenarios draws more and more attention. To solve the problem, one of the commonly used strategy is projecting the 3D point cloud into 2D (e.g., bird-eye-view \cite{chen2016monocular} or front-view \cite{wu2018squeezeseg}) to obtain the corresponding 2D detection result, then the final result can be obtained by re-projecting the 2D BBox into 3D. 

Benefiting from the development of graphics processing resources, volumetric convolutional approaches become another representative direction for 3D object detection. Voxelnet \cite{zhou2018voxelnet} is a pioneering method, which employs 3D convolution to detect the 3D objects by converting the LiDAR point cloud to voxels. Inspired by Voxelnet~\cite{zhou2018voxelnet}, SECOND \cite{yan2018second} and PointPillars \cite{lang2019pointpillars} have been proposed, which use different 3D voxel representations for 3D object detection. By using structure information of 3D point cloud, \cite{he2020structure} proposes a novel framework, which can improve the localization precision of single-stage detectors. Meanwhile, CenterPoint \cite{yin2021center} proposes to represent, detect, and track 3D objects as points, which detects centers and other attributes of the objects, then refines the estimated information using additional point features on the objects.

Meanwhile, PointNet \cite{qi2017pointnet} proposes a novel technique for point cloud feature extraction. % , which promotes the development of 3D object detection. 
Based on PointNet \cite{qi2017pointnet}, several state-of-the-art methods have been proposed for 3D object detection \cite{qi2018frustum,shi2019pointrcnn,shi2020points,shi2020pv}. To avoid destroying the hidden information about free space, \cite{hu2020you} proposes to utilize 3D ray casting and batch-based gradient learning strategies for 3D object detection. %To learn more discriminative point cloud features, Pv-rcnn \cite{shi2020pv} proposes a effective framework which consists of 3D voxel Convolutional Neural Network (CNN) and PointNet-based set abstraction, thus better 3D object detection results can be obtained. Part-A2 net \cite{shi2020points} extends PointRCNN to a novel and strong point-cloud-based 3D object detection framework, which consists of the part-aware stage and the part-aggregation stage and better performance can be achieved.

Besides, \cite{zhou2019iou} and \cite{yang2019std} propose effective intersection over-union (IoU) operations to generalize the losses in 3D object detection, which improve the accuracy of 3D object detection.

\subsection{Fusion-based 3D object detection}
Many approaches have been proposed for 3D object detection to leverage fusion strategy for better performance. MV3D \cite{chen2017multi} employs a compact multi-view representation to encode the sparse 3D point cloud, which is composed of 3D object proposal generation and multi-view feature fusion. Meanwhile, using LIDAR point cloud and RGB image as input, AVOD \cite{ku2018joint} proposes neural network architecture for 3D object detection, which consists of a region proposal network (RPN) and a second stage detector network. Besides, to solve the problem of localizing objects in point clouds of large-scale scenes, Frustum pointnets \cite{qi2018frustum} leverages both mature 2D object detectors and advanced 3D deep learning for 3D object detection with RGB-D data as input. \cite{zhou2020joint} proposes a simple but practical detection framework to jointly predict the 3D BBox and instance segmentation. 

Meanwhile, 3D maps always contain geographic, geometric and semantic priors that can improve the performance of many tasks. Using dense priors provided by large-scale crowd-sourced maps, \cite{wang2015holistic} proposes a novel framework to build a holistic model for multi-tasks of 3D object detection, semantic segmentation and depth reconstruction. Early attempt of fusing HDMaps, such as HDNet \cite{yang2018hdnet} shows that strong priors provided by HDMaps can boost the performance and robustness of modern 3D object detectors.

\begin{figure*}[ht!]
	\centering
	\includegraphics[width=0.85\textwidth]{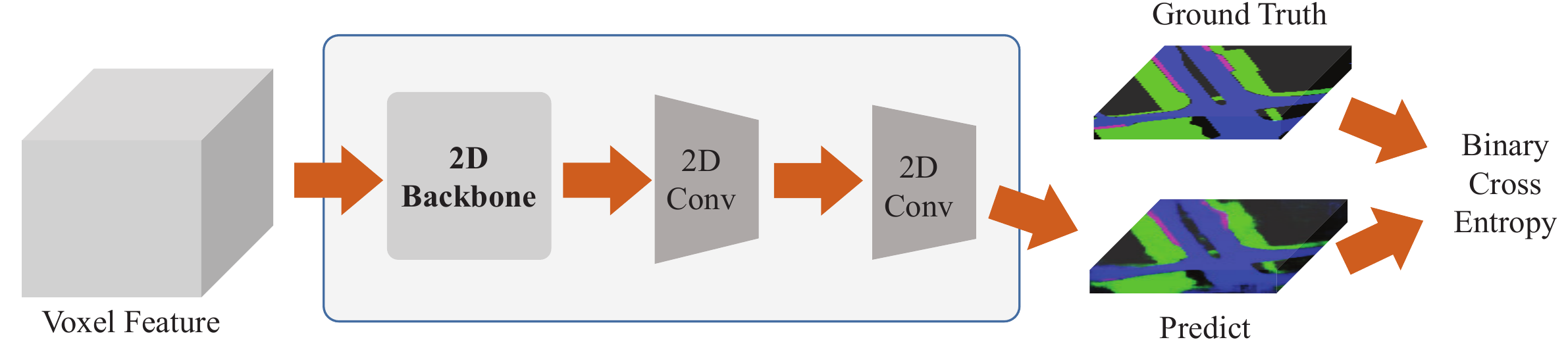}
	\centering
	\caption{The architecture of \textit{MapSeg} Module. Fill the voxel feature into the 2D backbone network and then follow two \textit{2D Conv} layers. Finally, the binary cross-entropy loss is evaluated between the predicted segmentation map and the ground truth map.}
	\label{Fig:map_seg}
\end{figure*}

%%%%%%%%%%%%%%%%%%%%%%%%%%%%%%%%%%%%%%%%%%%%%%
\section{Proposed Approach} \label{sec:method}
%%%%%%%%%%%%%%%%%%%%%%%%%%%%%%%%%%%%%%%%%%%%%%%%%

In this section, we present our MapFusion framework.

%%%%%%%%%%%%%%%%%%%%%%%%%%%%%%%%%%%%%%%%%
\subsection{Overview}
%%%%%%%%%%%%%%%%%%%%%%%%%%%%%%%%%%%%%%
The overview of the proposed framework is illustrated in Fig. \ref{Fig:Framework} which can be roughly divided into two parts: the standard 3D object detection block and the map feature extraction block. The standard LiDAR-based 3d object detection pipeline is described in the top red dotted box. The input LiDAR point cloud is sent to a 3d feature extractor such as 3D Sparse Convolution and outputs the features for voxels. The map feature extraction block is denoted as the blue dotted box, which takes the HDMap as the input. After a 2D feature extractor, the map feature with the same size of voxel feature is extracted. Then the 3D point cloud features and the map information are aggregated with a concatenation operation for each voxel.
%Then a 2d backbone network is employed for further feature aggregation for each voxel. 
Then, a detection head including the region proposals, box regression, and categories classification follows the fused features. In addition, an auxiliary segmentation head, namely \textit{MapSeg}, is added to further improve the feature extraction capability. Our MapFusion is an end-to-end framework that can be easily integrated into any standard 3d object detection pipeline with only slight modifications. Details of the MapFusion are introduced in the following subsections. In addition, the detailed network structure of \textit{2D Feature Extractor} and \textit{MapSeg}  are given in the Fig.\ref{Fig:2d_feature_extractor} and Fig.\ref{Fig:map_seg}.

%%%%%%%%%%%%%%%%%%%%%%%%%%%%%%%%%%%%%%%%%
\subsection{HDMap Representation}
%%%%%%%%%%%%%%%%%%%%%%%%%%%%%%%%%%%%%%

HDMaps contain rich information on road elements, such as drivable areas, walking areas, and lanes. % NuScenes dataset provides APIs to retrieve and query a certain record within the map layers. 
We use a raster representation by rendering the semantic elements with ego car at the center of the image.  % to the pose of ego vehicle. 
For the object detection tasks, only three kinds of elements are chosen here which are ``drivable\_area'', ``walkway'', and ``carpark\_area'' specifically. 

Instead of using the three raster images directly for fusion, we leverage a \textit{2D Feature Extractor} module to extract high-level features from the three raster images. The structure of \textit{2D Feature Extractor} is illustrated in Fig. \ref{Fig:2d_feature_extractor}, which is a stack of six similar layers including one 2D convolution with 3 × 3 kernels, batch
normalization, and a ReLU activate function. The filter number for the six layers are 16, 16, 32, 32, 32, 64, and 64 respectively. Specifically, we keep the image size unchanged before and after the \textit{2D Feature Extractor} block.

 %As shown in Fig. \ref{Fig:2d_feature_extractor}, \textit{the 2D Feature Extractor} is composed of a stack of N = 6 similar layers while only the filter number changes.   aggregate the neighbor information to expand the representation ability.

%rasterized, The resolution of map rasterization
%Map Feature Extractor

\subsection{FeatureAgg Module}

The \textit{FeatureAgg} module aims at fusing the extracted map features and the voxel features. For simplicity, we keep the voxel feature and the map feature with the same size and concatenate the two tensors along the feature channel. Although the operation is extremely simple, it gives satisfactory fusion results. In addition, we find that the performance can be further improved if a 1$\times$1 convolutional operation is added before sending the concatenated features into the next detection head. Further analysis can be found in the ablation study part.  

%In piratically, we have tried other strategies for fusing the two features for example 
%is about how to fusion the HDMap information with voxels. We let the Voxel feature tensor and Map feature tensor have the same H (height) and W (width) to simplify the subsequent alignment between voxels tensor and map tensor. 

\subsection{MapSeg Module}

\textit{MapSeg} module is an auxiliary segmentation head that takes the voxel feature as input and outputs the map segmentation predictions. The prediction is supervised by the ground-truth map image. The intention of this module is to learn the road structure information from the input point cloud directly. In fact, this kind of information is inherently learn-able because the derivable areas mostly have distinctive structures (e.g., flat) than Non-driving zones.   
%\textit{MapSeg} Module helps to improve the representation ability from the 3D Feature Extractor and 2D Backbone.

The architecture of \textit{MapSeg} is illustrated in Fig. \ref{Fig:map_seg}. Due to the possibility of overlap between different map elements definition, a binary cross-entropy loss is employed here for multi-label segmentation. The 2D Backbone used here is similar to the detection head. Following the 2D backbone, there are two additional \textit{2D Conv} layers which contain a 2D convolutional layer, batch normalization, and a ReLU operation. \textit{MapSeg} is only used during the training stage, it will be removed during inference process.

\begin{table*}[ht!]
\centering
\resizebox{0.99\textwidth}{!}
{
	\begin{tabular}{l|| c | c | c c c c c c c c c c}
	\hline
    \multicolumn{1}{c||}{\multirow{2}{*}{Methods}} &
    \multicolumn{1}{c|}{\multirow{2}{*}{\textbf{NDS}(\%)}} &
    \multicolumn{11}{c}{\textbf{AP} (\%)} \\
    
    \multicolumn{1}{c||}{} & \multicolumn{1}{c|}{} & \multicolumn{1}{l|}{mAP} & \multicolumn{1}{l}{Car} & \multicolumn{1}{l}{Pedestrian} & \multicolumn{1}{l}{Bus} & \multicolumn{1}{l}{Barrier} & \multicolumn{1}{l}{T.C.} &
    \multicolumn{1}{l}{Truck} & \multicolumn{1}{l}{Trailer} & \multicolumn{1}{l}{Moto.} & \multicolumn{1}{l}{Cons.} &
    \multicolumn{1}{l}{Bicycle} \\ \hline
    %\multicolumn{1}{l}{mAP} & \multicolumn{1}{l}{Car} & \multicolumn{1}{l}{Pedestrian} & \\
    %\textbf{Methods} & NDS & mAP & Car & Pedestrian & Bus & Barrier & T.C. & Truck & Trailer & Moto. & Cons. & Bicycle &  \hline

	SECOND \cite{yan2018second} (w/o MF) & 60.80 & 49.62 & 80.72 & 76.78 & 65.49 & 58.82 & 57.33 & 48.30 & 33.85 & 38.72 & 19.07 & 17.14 \\
	SECOND \cite{yan2018second} (w MF) &  62.04  & 50.89  &  81.83 & 77.83 & 67.71 & 59.80 & 58.19 & 50.31 & 37.85 & 40.04 & 17.68 & 17.65 \\
	Improvement $\uparrow$ & \textcolor{red}{+1.24} & \textcolor{red}{+1.27} & \textcolor{red}{+1.11} & \textcolor{red}{+1.05} & \textcolor{red}{+2.22} & \textcolor{red}{+0.98} & \textcolor{red}{+0.86} & \textcolor{red}{+2.01} & \textcolor{red}{+4.00} & \textcolor{red}{+1.32} & -1.39 & \textcolor{red}{+0.51} \\
    \hline
	
	PointPillars \cite{lang2019pointpillars} (w/o MF)&  57.45 & 43.87 & 81.10 & 70.91 &  62.69 & 47.11 & 45.04 & 49.36 & 35.53 & 30.34 & 11.41 & 5.20 \\
	PointPillars \cite{lang2019pointpillars} (w MF) & 58.95 & 46.66 & 81.21 & 72.22 & 65.51 & 54.19 & 52.43 & 45.21 & 37.79 & 36.74 & 14.34 & 6.92  \\
	Improvement $\uparrow$ & \textcolor{red}{+1.50} & \textcolor{red}{+2.79} & 
	\textcolor{red}{+0.11 } & \textcolor{red}{+1.31 } & \textcolor{red}{+2.82 } & \textcolor{red}{+7.08 } & \textcolor{red}{+7.39} & -4.15 & \textcolor{red}{+2.26 } & \textcolor{red}{+6.40 } &  \textcolor{red}{ +2.93}  & \textcolor{red}{+1.72 } \\
    \hline
	
	CenterPoint \cite{yin2021center} (w/o MF)&  67.13  & 59.43  &  85.97 & 85.46 & 68.50 & 68.20 & 69.43 & 58.25 & 38.81 & 59.57 & 19.21 & 40.94 \\
	CenterPoint \cite{yin2021center} (w MF) &  67.97 & 60.61  & 86.38 &  86.30 & 70.27 & 70.57 & 70.22 & 58.46 & 40.98 & 62.09 & 17.81 & 43.01\\
	Improvement $\uparrow$ & \textcolor{red}{+0.84} & \textcolor{red}{+1.18} & \textcolor{red}{+0.41} & \textcolor{red}{+0.84} & \textcolor{red}{+1.77} & \textcolor{red}{+2.37} & \textcolor{red}{+0.79} & \textcolor{red}{+0.21} & \textcolor{red}{+2.17} & \textcolor{red}{+2.52} & -1.40 & \textcolor{red}{+2.07} \\
    \hline

	%CenterPoint \cite{shi2020pv}  & 88.86 & 78.83  \\
	%CenterPoint \cite{shi2020pv} w MapFusion & 89.57 & 83.90 \\
	 \hline

	\end{tabular}
}
\caption{\normalfont Evaluation results of MapFusion on nuScenes validation dataset. NDS and mAP mean nuScenes detection score and mean Average Precision. MF is short for MapFusion. T.C., Moto. and Cons. are short for traffic cone, motorcycle, and construction vehicle, respectively. The improvements of each method are demonstrated in red, where ``w'' and ``w/o'' stand for ``with'' and ``without'' in short.}
\label{tab:eval_on_nuscenes}
\end{table*}

%The map image can be treated as a segmentation target for voxels in the meanwhile. 
\subsection{Data Augmentation}

Data augmentation is a commonly used strategy to increase data diversity and is extremely important for deep learning-based approaches for improving the model's generalization ability. 
For LiDAR-based object detection, commonly used data augmentation strategies include random rotation, flipping, and scaling. To align the features extracted from the point cloud and the HDMaps, for each sample, we keep the augmentation parameters are the same for both the two inputs.

%Our co-Augmentation module generates randomly augmentation parameters, and sends them to LiDAR frame feature extraction and HDMap feature extraction, while ensures the alignment between LiDAR point cloud and map.

%%%%%%%%%%%%%%%%%%%%%%%%%%%%%%%%%%%%%%%%%%%%%%
\section{Experimental Results} \label{sec:experiments}
%%%%%%%%%%%%%%%%%%%%%%%%%%%%%%%%%%%%%%%%%%%%%%%%%%%%%%

MapFusion is a general framework and can be easily integrated into mainstream 3d object detection methods. In this section, we first introduce the baseline detectors that are used to evaluate MapFusion. Then, more details of the used dataset and the experiment setups are given. Quantitative and qualitative results are demonstrated to prove the effectiveness of the proposed MapFusion. %Finally, by comparing with these baselines, we show the effectiveness of our method.

%MapFusion is a general framework and can be easily integrated into mainstream 3d object detection methods. In this section, we first introduce the baseline detectors we used to evaluate MapFusion. We then give more details about the dataset and our experiment setups. Finally, by comparing with these baselines, we show the effectiveness of our method.

\subsection{Baseline Detectors}
Three state-of-the-art point cloud-based 3d object detectors are compared here:
%\begin{enumerate}
    1) SECOND \cite{yan2018second} utilizes the sparse convolution to significantly increase the speed of both training and inference;
    2) PointPillars \cite{lang2019pointpillars} uses a pillar (vertical columns) representation and regards the pillars as a pseudo image. Then a standard 2D detection backbone can then be employed;
    3) CenterPoint \cite{yin2021center} is a strong anchor-free baseline, which ranks the top among all LiDAR-only method in public nuScenes \cite{nuscenes2019} and Waymo \cite{sun2020scalability} dataset.
%\end{enumerate}

%%%%%%%%%%%%%%%%%%%%%%%%%%%%%%%%%%
\subsection{Implementation Details}
%%%%%%%%%%%%%%%%%%%%%%%%%%%%%%%%%%

To compare fairly, the same experiment settings are used for the above baseline methods during training. The epoch number is set as 20, the optimization algorithm is AdamW \cite{ loshchilov2019decoupled} with a one-cycle learning rate policy, and the max learning rates for PointPillars, SECOND, and CenterPoint are 0.001, 0.001, and 0.003, while the weight decay is 0.01 and momentum is 0.9. The resolution of map rasterization we use here is $128 \times 128$.

Data augmentation is applied to guarantee data diversity and improve the robustness of the network. Our data augmentation strategies include 1) random rotation from [$-\frac{\pi}{4}, \frac{\pi}{4}$] around the gravity axis, 2) random flipping and 3) random scaling from the uniform distribution 0.95 to 1.05. Follow \cite{nuscenes2019}, 10 previous lidar sweeps are accumulated to leverage the temporal information.

Note that for CenterPoint\cite{yin2021center}, flip testing strategy and deformable convolution are employed to improve the performance as described by the original paper \cite{yin2020center}.

\subsection{nuScenes Dataset}

We evaluate our MapFusion framework on the challenging nuScenes 3D object detection benchmark~\cite{nuscenes2019}, since the KITTI benchmark~\cite{geiger2012we} does not provide HDMaps. nuScenes is a large-scale dataset with a total of 1,000 scenes, where 700 scenes (28,130 samples) are used for training, 150 scenes (6019 samples) are used for validation and 150 scenes (6008 samples) are used for testing. The samples (also named as keyframes) in each video are annotated every 0.5s with a full 360-degree view, and their point clouds are densified by the 10 non-keyframe sweep frames, yielding around {300,000} point clouds with 5-dim representation $(x, y, z, r, \Delta t)$, where $r$ is the reflectance and $\Delta t$ describes the time lag to the keyframe (ranging from 0s to 0.45s). Besides, nuScenes requires detecting objects  with full 3D boxes, attributes and velocities for 10 classes, including cars, pedestrians, buses, bicycles, etc.

For generating the raster images of HDMaps, NuScenes dataset provides APIs to retrieve and query a certain record, such as ``drivable\_area'', ``walkway'', and ``carpark\_area'' specifically within the map layers.

\subsection{Evaluation Metrics} \label{sec:metrics}

The official evaluation metrics are an average detection accuracy among all classes. For 3D detection, the main accuracy metric is mean Average Precision (mAP) \cite{everingham2010pascal} and nuScenes detection score (NDS). The mAP uses a bir-deye-view center distance $<$ 0.5m, 1m, 2m, 4m instead of standard 3D box IoU (Intersection-over-Union). NDS is a weighted average of mAP and other attributes metrics, including translation, scale, orientation, velocity, and other box attributes \cite{nuscenes2019}.

\subsection{Evaluation Results}
 
%In this section, we evaluate MapFusion on nuScenes dataset with the three baseline detectors we have introduced. The evaluation results are highlighted in Tab. \ref{tab:eval_on_nuscenes}. As the table shows, we can achieve xxxx point improvements for mean Average Precision (mAP).

We evaluate MapFusion on nuScenes dataset with the three baseline detectors which are introduced before, including SECOND, PointPillars and CenterPoint. Tab.~\ref{tab:eval_on_nuscenes} demonstrates the quantitative results with and without the proposed MapFusion (MF). As shown in Tab.~\ref{tab:eval_on_nuscenes}, the proposed Map Fusion (MF) can effectively improve the performance of the baselines in both nuScenes detection score (NDS) and mean Average Precision (mAP), which sufficiently proves the effectiveness of the proposed MapFusion. In specific, using the proposed MapFusion (MF), for NDS, the improvements of SECOND, PointPillars and CenterPoint are $1.24\%$, $1.50\%$ and $0.84\%$, while for mAP, the improvements of SECOND, PointPillars  and CenterPoint are $1.27\%$, $2.79\%$ and $1.18\%$, respectively. Meanwhile, as demonstrated in Tab.~\ref{tab:eval_on_nuscenes}, the proposed MapFusion can improve the performances for most classes, including small and large objects. Using MapFusion, for the commonly used small objects, such as ``Bicycle'', the improvements achieve $0.51\%$, $1.72\%$ and $2.07\%$ for the three baseline methods, and for large objects, such as ``Barrier'', the improvements achieve $0.98\%$, $7.08\%$ and $2.37\%$ for the three baseline methods. Note that for ``T.C.'', the average precision (AP) of PointPillars achieves $54.19\%$ with $7.39\%$ improvements. %can be improved we can achieve xxxx point improvements for mean Average Precision (mAP).

\section{Ablation Studies} \label{sec:ablation_study}

In this section, we explore the effectiveness and limitation of MapFusion. Firstly we show the impact of MapSeg branch and FeatureAgg module on the introduced 3d object baseline detectors, then the strategy for 2d feature extraction is further discussed. In the end, some visualization results are presented. All the experiments are evaluated on nuScenes validation dataset with the metrics defined in Sec. \ref{sec:metrics}.

\subsection{Influence of FeatureAgg and MapSeg}

As the most important components for MapFusion, an ablation experiment is conducted to explore the impact of FeatureAgg and MapSeg modules. The exact same baseline detectors, dataset, metrics and implementation details are used here following Sec. \ref{sec:experiments}. The results are shown in Tab. \ref{tab:ablation_study}, where we can clearly find the effectiveness of both FeatureAgg and MapSeg. With the help of MapSeg, the NDS and mAP value improved 0.32\% and 0.39\% averagely, respectively. In the meanwhile, FeatureAgg contributes 0.96\% and 1.02\% improvement averagely for NDS and mAP, respectively. The results also reveal that FeatureAgg has more impact on feature fusion than MapSeg, but the whole framework benefits the most by combining them.

\begin{table}[ht!]
\centering
	\resizebox{0.4\textwidth}{!}
	{%
		\begin{tabular}{l| c| cc }
			\hline
			%\multicolumn{1}{c }{\multirow{2}{*}{Methods}}& \multicolumn{3}{c}{\textbf{AP(IoU=0.7)}} \\
			%\multicolumn{1}{c}{}& \multicolumn{1}{l}{Easy} & \multicolumn{1}{l}{Mod} & \multicolumn{1}{l}{Hard}\\ \hline
			Methods & NDS & mAP \\ \hline
            SECOND baseline & 60.80 & 49.62  \\
            SECOND w MapSeg & 61.25 & 49.92 &\\
            SECOND w FeatureAgg & 61.46 & 50.19 & \\
            SECOND w MapFusion  & \textbf{62.04}  & \textbf{50.89} & \\ \hline

            PointPillar baseline &  57.45 & 43.87 \\
            PointPillar w MapSeg &  57.82  & 44.61 \\
            PointPillar w FeatureAgg & 58.69 & 45.64 & \\
            PointPillar w MapFusion  & \textbf{58.95} & \textbf{46.66} \\ \hline
            
            CenterPoint baseline & 67.13  & 59.43 \\
            CenterPoint w MapSeg & 67.27 & 59.56 &\\
            CenterPoint w FeatureAgg &  67.91 & 60.14 & \\
            CenterPoint w MapFusion  & \textbf{67.97} & \textbf{60.61}   \\ \hline
            
		\end{tabular}
	}
	\caption{\normalfont Evaluate the effectiveness of FeatureAgg and MapFusion on public nuScenes validation dataset, where ``w'' stands for ``with''.}
	\label{tab:ablation_study}
\end{table}

%\subsection{Early Fusion and Late Fusion}
%\subsection{Resolution of Maps}

\subsection{Influence of 2D Feature Extractor}

In this section, we explore the effectiveness of 2D Feature Extractor, two different settings are evaluated, including 1) \textit{Simple Fusion}, which directly use the original rendered map image as feature; 2) \textit{Deep Fusion}, which use the feature generated by the proposed architecture (2D Feature Extractor) in MapFusion. 

The evaluation results can be find in Tab \ref{tab:2d_feature}. \textit{Simple Fusion} with the map image directly also contribute to the final performance with 0.37\% and 0.74\% for NDS and mAP, respectively. But \textit{Deep Fusion}  outperforms the \textit{Simple Fusion} with 1.13\% and 2.05\% for for NDS and mAP, respectively. The results show that \textit{Deep Fusion} has better performance, benefits from the neighbor information aggregation by convolutions operation. The structure of proposed 2D Feature Extractor is very simple but effective, with a stronger network and pretrained on other dataset may get better performance. Since related technologies on 2d feature extraction have been widely researched, no further exploration of relevant issues will be addressed here. 

%2) \textit{Pretrained ResNet}, which is a strong ready-made model ResNet \cite{he2016deep} that has been pretrained on ImageNet \cite{deng2009imagenet}; 

\begin{table}[ht!]
\centering
	\resizebox{0.4\textwidth}{!}
	{%
		\begin{tabular}{l| c| cc }
			\hline
            Methods & NDS & mAP \\ \hline
            PointPillar baseline &  57.45 & 43.87 \\ 
            PointPillar w \textit{Simple Fusion}  & 57.82 & 44.61 \\ 
            %PointPillar w  &  & & \\ 
            PointPillar w \textit{Deep Fusion}  & \textbf{58.95} & \textbf{46.66} \\ \hline

		\end{tabular}
	}
	\caption{\normalfont Evaluate the effectiveness of 2D Feature Extractor on public nuScenes validata dataset with PointPillars~\cite{lang2019pointpillars}, where ``w'' stands for ``with''.}
	\label{tab:2d_feature}
\end{table}

\subsection{Different Feature Aggregation Methods}

In this section, we explore the impact of different feature aggregation methods, three strategies are used here, 1) \textit{Simple Concat}, which simply concatenate the voxel feature tensor and map feature tensor along the channel dimension; 2) \textit{Deep Concat v1}, which use a $1\times1$ convolutional operation after the feature concatenation, while the output channel number of the convolution layer is the same with voxel feature tensor; 3) \textit{Deep Concat v2}, which is similar with \textit{Deep Concat v1}, but the output channel number of the convolution layer keeps the same with input, which is our proposed method.

The evaluation results are shown in \ref{tab:feature_agg_method}, from where we can find that the $1\times1$ convolutional operation helps to improve the performance, especially about 1\% improvement for mAP, due to its better feature aggregation ability between two different features. Performance degrades with channel number decrease after the $1\times1$ convolution in 
\textit{Deep Concat v1} setting, which is normal after features compression.

\begin{table}[ht!]
\centering
	\resizebox{0.4\textwidth}{!}
	{%
		\begin{tabular}{l| c| cc }
			\hline
            Methods & NDS & mAP \\ \hline
            PointPillar baseline & 57.45 & 43.87 \\ \hline
            PointPillar w \textit{Simple Concat}  & 58.65  &  45.67 \\ 
            PointPillar w \textit{Deep Concat v1}  & 58.81  &  46.09 \\ 
            PointPillar w \textit{Deep Concat v2}  & \textbf{58.95} & \textbf{46.66} \\ \hline

		\end{tabular}
	}
	\caption{\normalfont Evaluate the impact of different feature aggregation method on public nuScenes validata dataset with PointPillars~\cite{lang2019pointpillars}, where ``w'' stand for ``with''.}
	\label{tab:feature_agg_method}
\end{table}

\begin{figure}[ht!]
	\centering
	\includegraphics[width=0.45\textwidth]{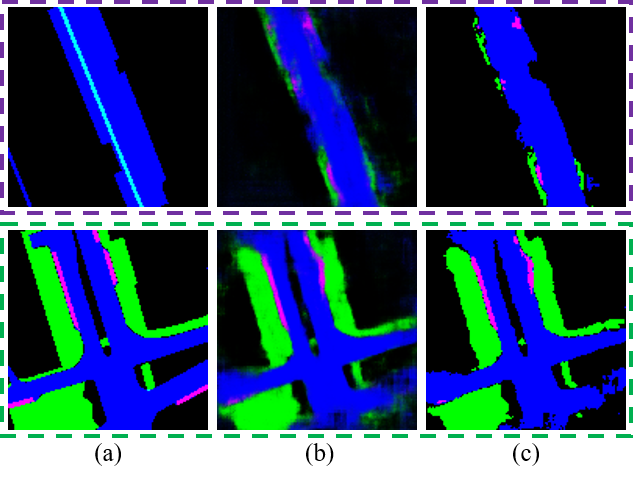}
	\centering
	%\caption{Visualization of segmentation results for the MapSeg module. (a) and (d) are the original map image (groundturth); (b) and (e) are the inference segmentation results of the corresponding voxels;}
	%\vspace{-0.1 in}
	\caption{Visualization of segmentation results for the MapSeg module. (a) shows the original map images (groundturth); (b) shows the inference segmentation results of the corresponding voxels, and (c) shows the final segmentation results of MapSeg module.}
	\label{Fig:map_seg_vis}
\end{figure}

\subsection{Visualization of MapSeg Results}

Fig.~\ref{Fig:map_seg_vis} demonstrates the segmentation results of the proposed MapSeg module, where (a) shows the original map images (groundtruth), (b) shows the inference segmentation results of the corresponding voxels and (c) shows the final segmentation results of MapSeg module. As shown in Fig.~\ref{Fig:map_seg_vis}, comparing with groundtruth, it is obvious to find that the proposed MapSeg module can effectively segment accurate results, which can boost the performance of the proposed approach.

\begin{figure}[ht!]
	\centering
	\includegraphics[width=0.45\textwidth]{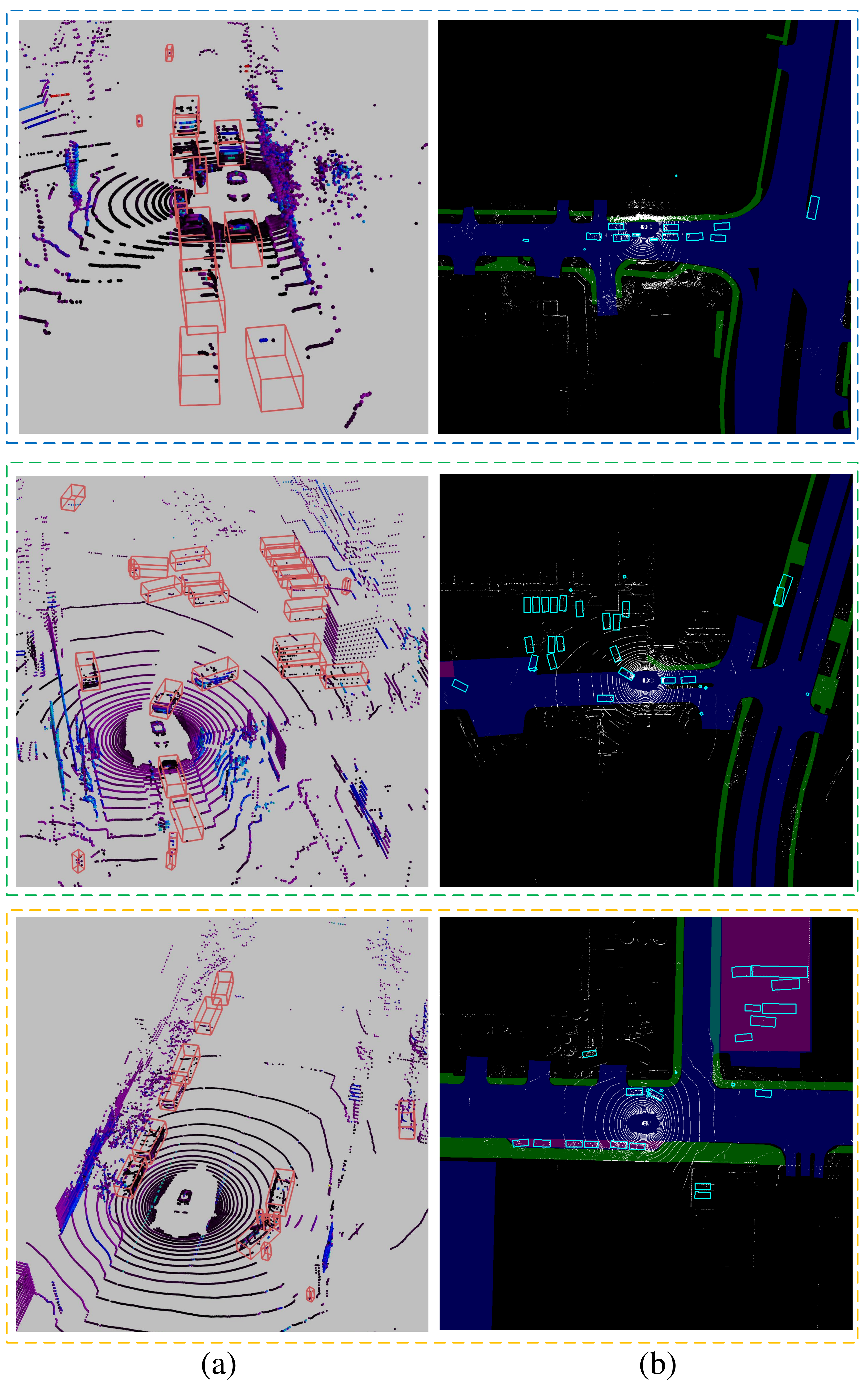}
	\centering
	%\caption{Visualization of segmentation results for the MapSeg module. (a) and (d) are the original map image (groundturth); (b) and (e) are the inference segmentation results of the corresponding voxels;}
	
	%\vspace{-0.1 in}
	
	\caption{Qualitative detection results of CenterPoint~\cite{yin2021center} with the proposed MapFusion, where (a) and (b) are displayed in front view and bird-eye-view with map respectively.}
	\label{Fig:det_vis}
\end{figure}

%\subsection{Limitations}

\subsection{ Detection Results Visualization}

%In Fig. \ref{Fig:det_vis} we demonstrate some detection results with MapFusion based on CenterPoint.

Fig. \ref{Fig:det_vis} demonstrates the quantitative detection results of CenterPoint~\cite{yin2021center} with the proposed MapFusion on the nuScenes ``val'' dataset. (a) and (b) illustrates the results of front view and bird-eye-view maps, respectively. We can see that CenterPoint with the proposed MapFusion can sufficiently detect all the 3D Bboxes of the objects, which proves the effectiveness of the proposed approach.

%%%%%%%%%%%%%%%%%%%%%%%%%%%%%%%%%%%%%
\section{Conclusion and Future Works}
%%%%%%%%%%%%%%%%%%%%%%%%%%%%%%%%%%%%%%%%

In this work, we present a map data fusion framework that can effectively explore the inherent relationship between road structure and the perception task. By fusing the map information, the 3D object detection performance is improved for all three different baseline methods. Qualitative analysis shows that false-positive detection can be largely suppressed. Currently, the proposed framework can fuse map information with LiDAR point cloud, and in the future, we plan to extend our framework to fuse the data from other sensors such as radar, camera, etc with HDMaps. In addition, as HDmaps might not be available everywhere, we plan to develop map prediction methods and integrate widely accessible regular maps with the MapFusion framework. 
% In addition, all the information inputhodss are loosely decoupled, and make the framework has the flexibility to add or remove any sensor while the framework can still function well. 

\bibliographystyle{IEEEtran}
\bibliography{IEEEabrv,ref}

\begin{thebibliography}{10}
\providecommand{\url}[1]{#1}
\csname url@rmstyle\endcsname
\providecommand{\newblock}{\relax}
\providecommand{\bibinfo}[2]{#2}
\providecommand\BIBentrySTDinterwordspacing{\spaceskip=0pt\relax}
\providecommand\BIBentryALTinterwordstretchfactor{4}
\providecommand\BIBentryALTinterwordspacing{\spaceskip=\fontdimen2\font plus
\BIBentryALTinterwordstretchfactor\fontdimen3\font minus
  \fontdimen4\font\relax}
\providecommand\BIBforeignlanguage[2]{{%
\expandafter\ifx\csname l@#1\endcsname\relax
\typeout{** WARNING: IEEEtran.bst: No hyphenation pattern has been}%
\typeout{** loaded for the language `#1'. Using the pattern for}%
\typeout{** the default language instead.}%
\else
\language=\csname l@#1\endcsname
\fi
#2}}

\bibitem{Takeda2020Survey}
E.~{Yurtsever}, J.~{Lambert}, A.~{Carballo}, and K.~{Takeda}, ``A survey of
  autonomous driving: Common practices and emerging technologies,'' \emph{IEEE
  Access}, vol.~8, pp. 58\,443--58\,469, 2020.

\bibitem{arnold2019survey}
E.~Arnold, O.~Y. Al-Jarrah, M.~Dianati, S.~Fallah, D.~Oxtoby, and
  A.~Mouzakitis, ``A survey on 3d object detection methods for autonomous
  driving applications,'' \emph{IEEE Transactions on Intelligent Transportation
  Systems}, vol.~20, no.~10, pp. 3782--3795, 2019.

\bibitem{ren2015faster}
S.~Ren, K.~He, R.~Girshick, and J.~Sun, ``Faster r-cnn: Towards real-time
  object detection with region proposal networks,'' in \emph{Advances in neural
  information processing systems}, 2015, pp. 91--99.

\bibitem{liu2016ssd}
W.~Liu, D.~Anguelov, D.~Erhan, C.~Szegedy, S.~Reed, C.-Y. Fu, and A.~C. Berg,
  ``Ssd: Single shot multibox detector,'' in \emph{European conference on
  computer vision}.\hskip 1em plus 0.5em minus 0.4em\relax Springer, 2016, pp.
  21--37.

\bibitem{lang2019pointpillars}
A.~H. Lang, S.~Vora, H.~Caesar, L.~Zhou, J.~Yang, and O.~Beijbom,
  ``Pointpillars: Fast encoders for object detection from point clouds,'' in
  \emph{Proceedings of the IEEE Conference on Computer Vision and Pattern
  Recognition}, 2019, pp. 12\,697--12\,705.

\bibitem{shi2019pointrcnn}
S.~Shi, X.~Wang, and H.~Li, ``Pointrcnn: 3d object proposal generation and
  detection from point cloud,'' in \emph{CVPR}, 2019.

\bibitem{zhou2020joint}
D.~Zhou, J.~Fang, X.~Song, L.~Liu, J.~Yin, Y.~Dai, H.~Li, and R.~Yang, ``Joint
  3d instance segmentation and object detection for autonomous driving,'' in
  \emph{Proceedings of the IEEE/CVF Conference on Computer Vision and Pattern
  Recognition}, 2020, pp. 1839--1849.

\bibitem{yan2018second}
Y.~Yan, Y.~Mao, and B.~Li, ``Second: Sparsely embedded convolutional
  detection,'' \emph{Sensors}, vol.~18, no.~10, p. 3337, 2018.

\bibitem{liang2019multi}
M.~Liang, B.~Yang, Y.~Chen, R.~Hu, and R.~Urtasun, ``Multi-task multi-sensor
  fusion for 3d object detection,'' in \emph{Proceedings of the IEEE/CVF
  Conference on Computer Vision and Pattern Recognition}, 2019, pp. 7345--7353.

\bibitem{wang2021rodnet}
Y.~Wang, Z.~Jiang, X.~Gao, J.-N. Hwang, G.~Xing, and H.~Liu, ``Rodnet: Radar
  object detection using cross-modal supervision,'' in \emph{Proceedings of the
  IEEE/CVF Winter Conference on Applications of Computer Vision}, 2021, pp.
  504--513.

\bibitem{yang2018hdnet}
B.~Yang, M.~Liang, and R.~Urtasun, ``Hdnet: Exploiting hd maps for 3d object
  detection,'' in \emph{Conference on Robot Learning}.\hskip 1em plus 0.5em
  minus 0.4em\relax PMLR, 2018, pp. 146--155.

\bibitem{yin2020center}
T.~Yin, X.~Zhou, and P.~Kr{\"a}henb{\"u}hl, ``Center-based 3d object detection
  and tracking,'' \emph{arXiv preprint arXiv:2006.11275}, 2020.

\bibitem{nuscenes2019}
H.~Caesar, V.~Bankiti, A.~H. Lang, S.~Vora, V.~E. Liong, Q.~Xu, A.~Krishnan,
  Y.~Pan, G.~Baldan, and O.~Beijbom, ``nuscenes: A multimodal dataset for
  autonomous driving,'' \emph{arXiv preprint arXiv:1903.11027}, 2019.

\bibitem{chen2016monocular}
X.~Chen, K.~Kundu, Z.~Zhang, H.~Ma, S.~Fidler, and R.~Urtasun, ``Monocular 3d
  object detection for autonomous driving,'' in \emph{Proceedings of the IEEE
  Conference on Computer Vision and Pattern Recognition}, 2016, pp. 2147--2156.

\bibitem{wu2018squeezeseg}
B.~Wu, A.~Wan, X.~Yue, and K.~Keutzer, ``Squeezeseg: Convolutional neural nets
  with recurrent crf for real-time road-object segmentation from 3d lidar point
  cloud,'' in \emph{2018 IEEE International Conference on Robotics and
  Automation (ICRA)}.\hskip 1em plus 0.5em minus 0.4em\relax IEEE, 2018, pp.
  1887--1893.

\bibitem{zhou2018voxelnet}
Y.~Zhou and O.~Tuzel, ``Voxelnet: End-to-end learning for point cloud based 3d
  object detection,'' in \emph{Proceedings of the IEEE Conference on Computer
  Vision and Pattern Recognition}, 2018, pp. 4490--4499.

\bibitem{he2020structure}
C.~He, H.~Zeng, J.~Huang, X.-S. Hua, and L.~Zhang, ``Structure aware
  single-stage 3d object detection from point cloud,'' in \emph{Proceedings of
  the IEEE/CVF Conference on Computer Vision and Pattern Recognition}, 2020,
  pp. 11\,873--11\,882.

\bibitem{yin2021center}
T.~Yin, X.~Zhou, and P.~Kr{\"a}henb{\"u}hl, ``Center-based 3d object detection
  and tracking,'' \emph{CVPR}, 2021.

\bibitem{qi2017pointnet}
C.~R. Qi, H.~Su, K.~Mo, and L.~J. Guibas, ``Pointnet: Deep learning on point
  sets for 3d classification and segmentation,'' in \emph{Proceedings of the
  IEEE Conference on Computer Vision and Pattern Recognition}, 2017, pp.
  652--660.

\bibitem{qi2018frustum}
C.~R. Qi, W.~Liu, C.~Wu, H.~Su, and L.~J. Guibas, ``Frustum pointnets for 3d
  object detection from rgb-d data,'' in \emph{Proceedings of the IEEE
  Conference on Computer Vision and Pattern Recognition}, 2018, pp. 918--927.

\bibitem{shi2020points}
S.~Shi, Z.~Wang, J.~Shi, X.~Wang, and H.~Li, ``From points to parts: 3d object
  detection from point cloud with part-aware and part-aggregation network,''
  \emph{IEEE Transactions on Pattern Analysis and Machine Intelligence}, 2020.

\bibitem{shi2020pv}
S.~Shi, C.~Guo, L.~Jiang, Z.~Wang, J.~Shi, X.~Wang, and H.~Li, ``Pv-rcnn:
  Point-voxel feature set abstraction for 3d object detection,'' in
  \emph{Proceedings of the IEEE/CVF Conference on Computer Vision and Pattern
  Recognition}, 2020, pp. 10\,529--10\,538.

\bibitem{hu2020you}
P.~Hu, J.~Ziglar, D.~Held, and D.~Ramanan, ``What you see is what you get:
  Exploiting visibility for 3d object detection,'' in \emph{Proceedings of the
  IEEE/CVF Conference on Computer Vision and Pattern Recognition}, 2020, pp.
  11\,001--11\,009.

\bibitem{zhou2019iou}
D.~Zhou, J.~Fang, X.~Song, C.~Guan, J.~Yin, Y.~Dai, and R.~Yang, ``Iou loss for
  2d/3d object detection,'' in \emph{2019 International Conference on 3D Vision
  (3DV)}.\hskip 1em plus 0.5em minus 0.4em\relax IEEE, 2019, pp. 85--94.

\bibitem{yang2019std}
Z.~Yang, Y.~Sun, S.~Liu, X.~Shen, and J.~Jia, ``Std: Sparse-to-dense 3d object
  detector for point cloud,'' in \emph{Proceedings of the IEEE International
  Conference on Computer Vision}, 2019, pp. 1951--1960.

\bibitem{chen2017multi}
X.~Chen, H.~Ma, J.~Wan, B.~Li, and T.~Xia, ``Multi-view 3d object detection
  network for autonomous driving,'' in \emph{Proceedings of the IEEE Conference
  on Computer Vision and Pattern Recognition}, 2017, pp. 1907--1915.

\bibitem{ku2018joint}
J.~Ku, M.~Mozifian, J.~Lee, A.~Harakeh, and S.~Waslander, ``Joint 3d proposal
  generation and object detection from view aggregation,'' \emph{IROS}, 2018.

\bibitem{wang2015holistic}
S.~Wang, S.~Fidler, and R.~Urtasun, ``Holistic 3d scene understanding from a
  single geo-tagged image,'' in \emph{Proceedings of the IEEE Conference on
  Computer Vision and Pattern Recognition}, 2015, pp. 3964--3972.

\bibitem{sun2020scalability}
P.~Sun, H.~Kretzschmar, X.~Dotiwalla, A.~Chouard, V.~Patnaik, P.~Tsui, J.~Guo,
  Y.~Zhou, Y.~Chai, B.~Caine, \emph{et~al.}, ``Scalability in perception for
  autonomous driving: Waymo open dataset,'' in \emph{Proceedings of the
  IEEE/CVF Conference on Computer Vision and Pattern Recognition}, 2020, pp.
  2446--2454.

\bibitem{loshchilov2019decoupled}
\BIBentryALTinterwordspacing
I.~Loshchilov and F.~Hutter, ``Decoupled weight decay regularization,'' in
  \emph{International Conference on Learning Representations}, 2019. [Online].
  Available: \url{https://openreview.net/forum?id=Bkg6RiCqY7}
\BIBentrySTDinterwordspacing

\bibitem{geiger2012we}
A.~Geiger, P.~Lenz, and R.~Urtasun, ``Are we ready for autonomous driving? the
  kitti vision benchmark suite,'' in \emph{2012 IEEE Conference on Computer
  Vision and Pattern Recognition}.\hskip 1em plus 0.5em minus 0.4em\relax IEEE,
  2012, pp. 3354--3361.

\bibitem{everingham2010pascal}
M.~Everingham, L.~Van~Gool, C.~K. Williams, J.~Winn, and A.~Zisserman, ``The
  pascal visual object classes (voc) challenge,'' \emph{International journal
  of computer vision}, vol.~88, no.~2, pp. 303--338, 2010.

\end{thebibliography}

\end{document}